\documentclass[letterpaper]{article} 
\usepackage{aaai2026}  
\usepackage{times}  
\usepackage{helvet}  
\usepackage{courier}  
\usepackage[hyphens]{url}  
\usepackage{graphicx} 
\usepackage{natbib}  
\usepackage{caption} 
\usepackage{algorithm}
\usepackage{algorithmic}

\usepackage{makecell}
\usepackage{multirow}
\usepackage{amssymb}
\usepackage{enumitem}
\usepackage{needspace}
\usepackage{newfloat}
\usepackage{listings}
\DeclareCaptionStyle{ruled}{labelfont=normalfont,labelsep=colon,strut=off} 
\floatstyle{ruled}
\newfloat{listing}{tb}{lst}{}
\floatname{listing}{Listing}
\title{Mono3DVG-EnSD: Enhanced Spatial-aware and Dimension-decoupled Text Encoding for Monocular 3D Visual Grounding}
\author{
    Yuzhen Li\textsuperscript{\rm 1},
    Min Liu\textsuperscript{\rm 1}\thanks{Corresponding Author},
    Zhaoyang Li\textsuperscript{\rm 1},
    Yuan Bian\textsuperscript{\rm 1},
    Xueping Wang\textsuperscript{\rm 2},
    Erbo Zhai\textsuperscript{\rm 1},
    Yaonan Wang\textsuperscript{\rm 1}
}
\affiliations{
    \textsuperscript{\rm 1}School of Artificial
Intelligence and Robotics, Hunan University, Changsha, Hunan, China\\
    \textsuperscript{\rm 2}College of Information
Science and Engineering, Hunan Normal University, Changsha, Hunan, China\\

    \{zzrs, liu\_min, zhaoyli, yuanbian, wang\_xueping, zhaierbo, yaonan\}@hnu.edu.cn
}

\usepackage{bibentry}

\begin{document}

\maketitle

\begin{abstract}
Monocular 3D Visual Grounding (Mono3DVG) is an emerging task that locates 3D objects in RGB images using text descriptions with geometric cues. However, existing methods face two key limitations. Firstly, they often over-rely on high-certainty keywords that explicitly identify the target object while neglecting critical spatial descriptions. Secondly, generalized textual features contain both 2D and 3D descriptive information, thereby capturing an additional dimension of details compared to singular 2D or 3D visual features. This characteristic leads to cross-dimensional interference when refining visual features under text guidance. To overcome these challenges, we propose Mono3DVG-EnSD, a novel framework that integrates two key components: the CLIP-Guided Lexical Certainty Adapter (CLIP-LCA) and the Dimension-Decoupled Module (D2M). The CLIP-LCA dynamically masks high-certainty keywords while retaining low-certainty implicit spatial descriptions, thereby forcing the model to develop a deeper understanding of spatial relationships in captions for object localization. Meanwhile, the D2M decouples dimension-specific (2D/3D) textual features from generalized textual features to guide corresponding visual features at same dimension, which mitigates cross-dimensional interference by ensuring dimensionally-consistent cross-modal interactions. Through comprehensive comparisons and ablation studies on the Mono3DRefer dataset, our method achieves state-of-the-art (SOTA) performance across all metrics. Notably, it improves the challenging Far(Acc@0.5) scenario by a significant +13.54\%.
\end{abstract}

%

\section{Introduction}

The ability to locate objects through linguistic instructions constitutes a fundamental capability for human-robot interaction systems. While 2D visual grounding methods \cite{huang2022deconfounded, zheng2025look, dai2025multi, xie2025phrase, linhuixiao24} have achieved significant progress in image understanding, their inherent limitation in depth perception restricts their capacity to interpret spatial relationships. Therefore, researchers have adopted multi-modal fusion techniques, which integrate complementary sensor data to enhance robust scene perception. For instance, RGB-D methods \cite{zehantan24,chendz20,achlioptasp20} are prevalent in indoor scene understanding and LiDAR-camera fusion \cite{linz25,jinyuanli25,shutinghe24} is widely used in outdoor robotic perception. Despite their effectiveness, the broader adoption of these methods is still limited by the high expenses of RGB-D and LiDAR sensors.

\begin{figure}[t]
\centering
\includegraphics[width=0.9\columnwidth]{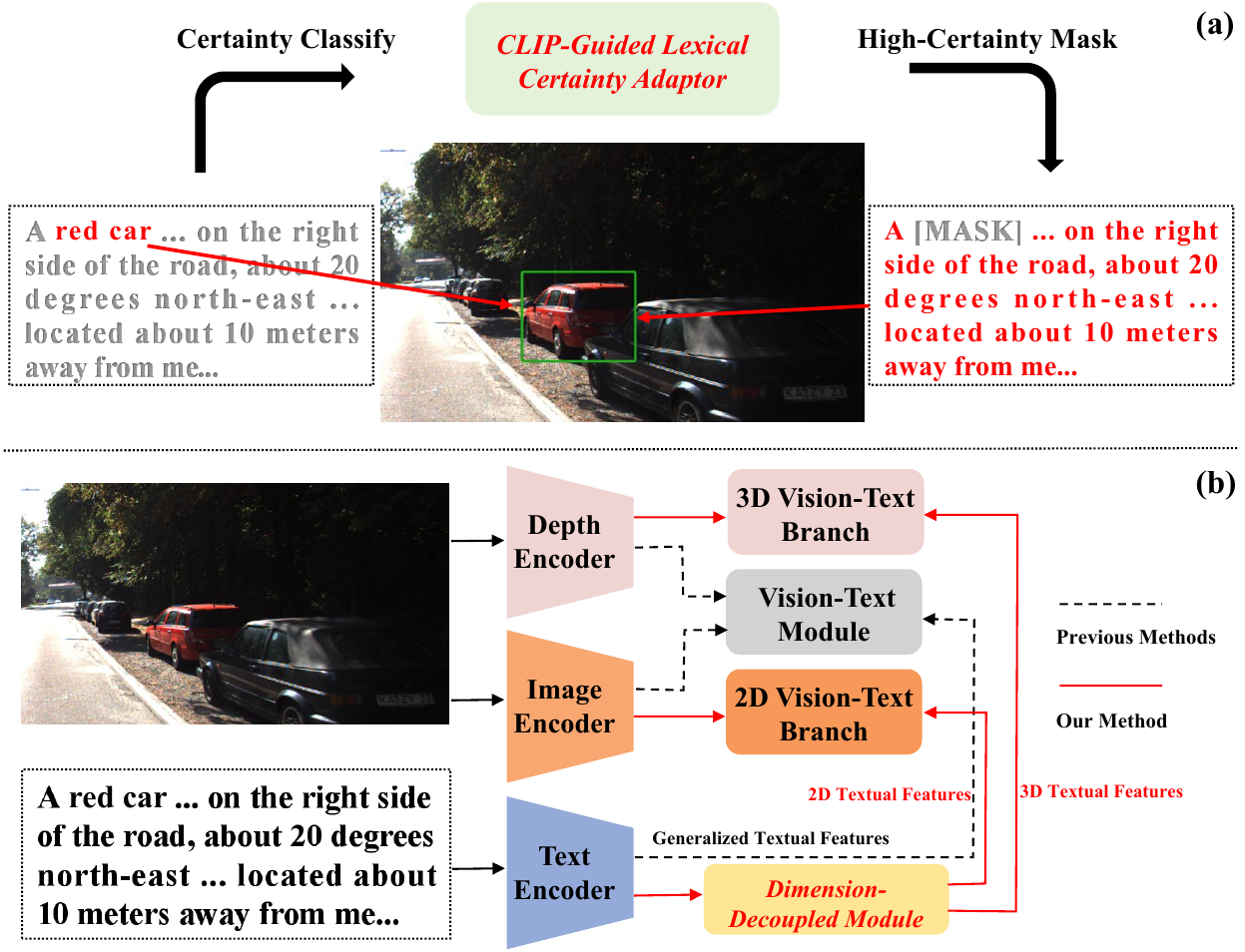} 
\caption{(a) Existing methods over-rely on high-certainty words (e.g., ``red car'') within descriptions (left), causing spatial description neglect. Our CLIP-LCA dynamically masks such high-certainty words during training, forcing the text encoder to comprehend spatial descriptors; (b) Previous methods employ generalized textual features (containing both 2D and 3D information) to refine both 3D visual (depth) features and 2D visual features, causing cross-dimensional interference. We propose D2M to decouple dimension-specific text features (2D and 3D separately) for dimensionally-consistent cross-modal interactions.}
\label{fig1}
\end{figure}

Monocular 3D object detection \cite{brazil2023omni3d, huang2022monodtr, li2024monolss} enables the estimation of 3D spatial location of objects from a single image. However,  it often fails to capture the semantic context of the 3D environment. This limits its applications in human interaction systems, like robotics, where understanding natural language instructions is crucial for instruction-guided object localization. To bridge this gap, researchers have introduced monocular 3D visual grounding (mono3DVG) \cite{zhan2024mono3dvg}, a novel task that uses linguistic captions with geometry information to accurately localize 3D objects from a single RGB image. Unlike traditional 3D visual grounding methods that rely on point clouds or depth maps as 3D representations, the monocular approach operates solely on RGB images. To compensate for the absence of explicit 3D geometry in visual data, the mono3DVG task supplements captions with geometry descriptions (e.g., ``10 meters away'').

Compared to traditional 3D visual grounding, mono3DVG offers a more efficient and economical alternative. However, existing mono3DVG methods face two critical challenges. Firstly, captions in the mono3DVG not only provide enriched 3D spatial descriptors but also contain high-certainty keywords whose identification directly enables precise target localization. This leads the model to rely on simple keyword matching rather than developing a robust understanding of the underlying spatial semantics, thereby compromising its generalization capability in complex scenarios. As illustrated in Fig.\ref{fig1} (a), the keyword ``red car'' in the text description is sufficient to accurately locate the corresponding object in the image. This phenomenon may lead the text encoder to over-rely on high-certainty words while neglecting critical spatial descriptors such as ``on the right side'', ``20 degrees north-east'' and ``10 meters away''. Consequently, the model exhibits poor performance in scenarios where precise object localization requires a comprehensive understanding of spatial relationships. Secondly, previous methods suffer from cross-dimensional interference caused by the interaction between textual and visual features, as presented in Fig.\ref{fig1} (b). These methods employ separate image and depth encoders to extract 2D and 3D visual (depth) features respectively, while the text encoder integrates 2D descriptions, 3D descriptions, and common semantic information from captions into generalized textual features. The generalized textual features typically contain multi-dimensional semantic information, whereas visual features extracted from image encoder or depth encoder capture only single-dimensional details. When employing generalized textual features to refine 2D or 3D visual features, the textual features with multi-dimensional information may introduce irrelevant dimensional noise into single-dimensional visual features. For example, 2D attributes in generalized textual features (e.g., ``red'', ``shirt'') can distort 3D visual feature encoding, leading to inaccurate depth estimation.

To address the issue of high-certainty words in captions that explicitly identify corresponding objects, we propose a CLIP-Guided Lexical Certainty Adapter (CLIP-LCA), which dynamically adjusts the certainty levels of these keywords to encourage the text encoder to capture spatial information. During the training phase, we calculate a similarity score between each word in the textual description and the corresponding target region by leveraging CLIP's \cite{radford2021learning} image-text representation alignment capabilities. Based on these scores, we categorize the words into high-certainty and low-certainty classes. Words identified as high-certainty are masked during training, which balances attention of the text encoder between high-certainty keywords and low-certainty spatial descriptors. CLIP-LCA enhances the text encoder's ability to understand spatial relationships in descriptions, enabling more comprehensive sentence comprehension and consequently improving localization accuracy.

To resolve the cross-dimensional interference problem, we propose a Dimension-Decoupled Module (D2M). The D2M framework first employs two parallel cross-attention modules with distinct learnable embeddings to decouple generalized textual features into dimension-specific representations. Specifically, a 2D learnable embedding interacts with the generalized textual features to capture coarse 2D textual features, while a parallel 3D learnable embedding simultaneously learns coarse 3D textual features. To further enhance the dimension-specific information of the two coarse textual features without additional data support, we propose a dual-branch reverse cross-attention module. In the 2D branch, the coarse 2D textual features serve as queries (Q) and values (V), while the coarse 3D textual features act as keys (K). The low-attention regions between Q and K indicate dimensional discrepancies, thereby identifying 2D-specific details that require enhancement. The enhancement process first inverts low-attention weights and then applies them through matrix multiplication to the V. The 3D branch operates symmetrically when reversing the qkv configuration to enhance 3D part. D2M generates refined 2D-specific and 3D-specific textual features that provide more precise dimension-specific guidance for corresponding dimensional visual features.

The primary contributions of this paper can be summarized as follows:
\begin{itemize}
    \item We propose the CLIP-Guided Lexical Certainty Adapter (CLIP-LCA), which enhances text encoder understanding of spatial relationships by dynamically adjusting word certainty. This mitigates over-reliance on high-certainty keywords while improving attention to spatial descriptors.
    \item We introduce the Dimension-Decoupled Module (D2M) that resolves cross-dimensional interference by employing disentangled 2D and 3D textual features to guide corresponding visual features, enabling dimensionally-consistent cross-modal refinement.
    \item Experimental results demonstrate that our Mono3DVG-EnSD achieves state-of-the-art performance across all metrics, with a significant +13.54\% improvement on the challenging Far(Acc@0.5) metric compared to previous methods.
\end{itemize}

\section{Related Work}

\subsection{Monocular 3D Object Detection}
Monocular 3D object detection focuses on predicting 3D bounding boxes from a single image and can be divided into two approaches depending on the use of supplementary data. One group of methods operates exclusively on monocular images, like M3D-RPN \cite{brazilg19}, which introduces a 3D region proposal network and utilizes depth-aware convolutions. Methods like SMOKE \cite{liuz20} and FCOS3D \cite{wangt21} adopt a key-point estimation strategy inspired by CenterNet \cite{zhoux19}, where size and location of 3D boxes are derived from heatmap peaks. To address the challenge of feature interference, MonoLSS \cite{zli24} proposes a Learnable Sample Selection (LSS) module that dynamically filters noise features, thereby improving 3D representation learning. MonoPair \cite{yongjianchen20} improves localization accuracy by modeling spatial relationships between neighboring objects, which is beneficial for occluded instances. MonoEF \cite{yunsongzhou21} introduces a novel approach to estimate camera extrinsics using vanishing points and horizon line detection, followed by a feature rectification module to correct distortions in latent representations. MonoCon \cite{tianfuwu22} employs an auxiliary learning strategy during training, where monocular contextual features are aligned with 3D bounding box properties, and the auxiliary head is discarded during inference to maintain computational efficiency. MonoDDE \cite{zhuolingli22} utilizes depth cues to generate multiple depth hypotheses per object. Another line of research enhances 3D detection by integrating auxiliary data like depth maps, point clouds, or CAD models. D4LCN \cite{mingyuding20} proposes depth-aware convolutions that adapt receptive fields according to inferred depth. DID-M3D \cite{liangpeng22} improves depth estimation by separating instance-specific depths into attribute and visual components. ROI-10D \cite{fabianm19} predicts 3D boxes via dense depth estimation. For LIDAR-based methods, CaDDN \cite{codyreading21} projects LIDAR-derived depth maps into a monocular network, then converts features into a bird's-eye-view space for detection. CMKD \cite{yuhong22} transfers knowledge across modalities, specifically distilling LIDAR-based features into visual representations. Beyond depth maps and LIDAR data, AutoShape \cite{zongdailiu21} utilizes keypoints extracted from CAD models to address limitations caused by sparse supervision.

\subsection{2D Visual Grounding}
Visual grounding, a task building upon object detection, seeks to create accurate correspondences between textual descriptions and specific areas in images. Initial methodologies in this domain primarily employed two-stage pipelines \cite{daqingliu19,sibeiyang19,lichengyu18}, dividing the process into distinct phases: region proposal generation and cross-modal alignment. In the first phase, an independent object detector (e.g., Faster R-CNN) produces candidate regions without considering textual context, which may lead to semantic inconsistencies due to the absence of language-aware guidance. To mitigate cross-modal semantic misalignment, MattNet \cite{lichengyu18} parses textual descriptions into structured elements to enable detailed vision-language interaction. In contrast to region-based methods, proposal-free techniques \cite{xinpengchen18,yueliao20,zhenyuanyang19} employ spatial feature fusion, allowing direct region prediction through thorough multimodal understanding. FAOA \cite{zhenyuanyang19} exemplifies this approach, combining fused visual-textual representations with YOLOv3 \cite{yolov3} for end-to-end visual grounding. Building on success of transformers across multiple domains, several transformer-based architectures have been adapted for visual grounding. The first attempt to leverage transformers in this task was proposed by TransVG \cite{dengj21}. Referring Transformer \cite{zhuolingli22} employs phrase-based queries to jointly perform region localization and segmentation in a unified framework. Many transformer-based methods \cite{zheng2025look, liuimproving, liu2025fedadamw} achieved success.

\subsection{3D Visual Grounding}
The establishment of benchmark datasets for 3D visual grounding began with the introduction of Referit3D \cite{achlioptasp20} and ScanRefer \cite{chendz20}. Earlier works followed a two-stage pipeline similar to 2D approaches. PointNet++ \cite{qic17} relies on a pre-trained detector for proposal generation and feature extraction. To enrich semantic understanding, SAT \cite{yangz2021} incorporates 2D object semantics for improved model training. For handling complex descriptions and localizing objects in point clouds, Feng et al. \cite{fengm21} propose three specialized modules. Recent advances in 3D visual grounding leverage transformer-based architectures, including 3DVG-Trans \cite{zhaol21}, LanguageRefer \cite{rohj22}, and Multi-View Trans \cite{huangs22}. Unified frameworks [5][6] have been proposed to jointly address visual grounding and dense captioning within a single model. Liu et al. \cite{liuh21} proposed a novel 3D visual grounding task using RGB-D images. Unlike prior indoor-focused studies that primarily predicted furniture objects, Lin et al. \cite{linz25} extended the scenario to outdoor environments by incorporating both 2D images and 3D point clouds. The practical deployment of LiDAR and RGB-D based systems faces challenges due to hardware cost and availability constraints. As a solution, Mono3DVG \cite{zhany24} establishes a monocular alternative that combines geometrically detailed text with single RGB images to predict 3D bounding boxes.

\begin{figure*}[t]
\centering
\includegraphics[width=0.9\textwidth]{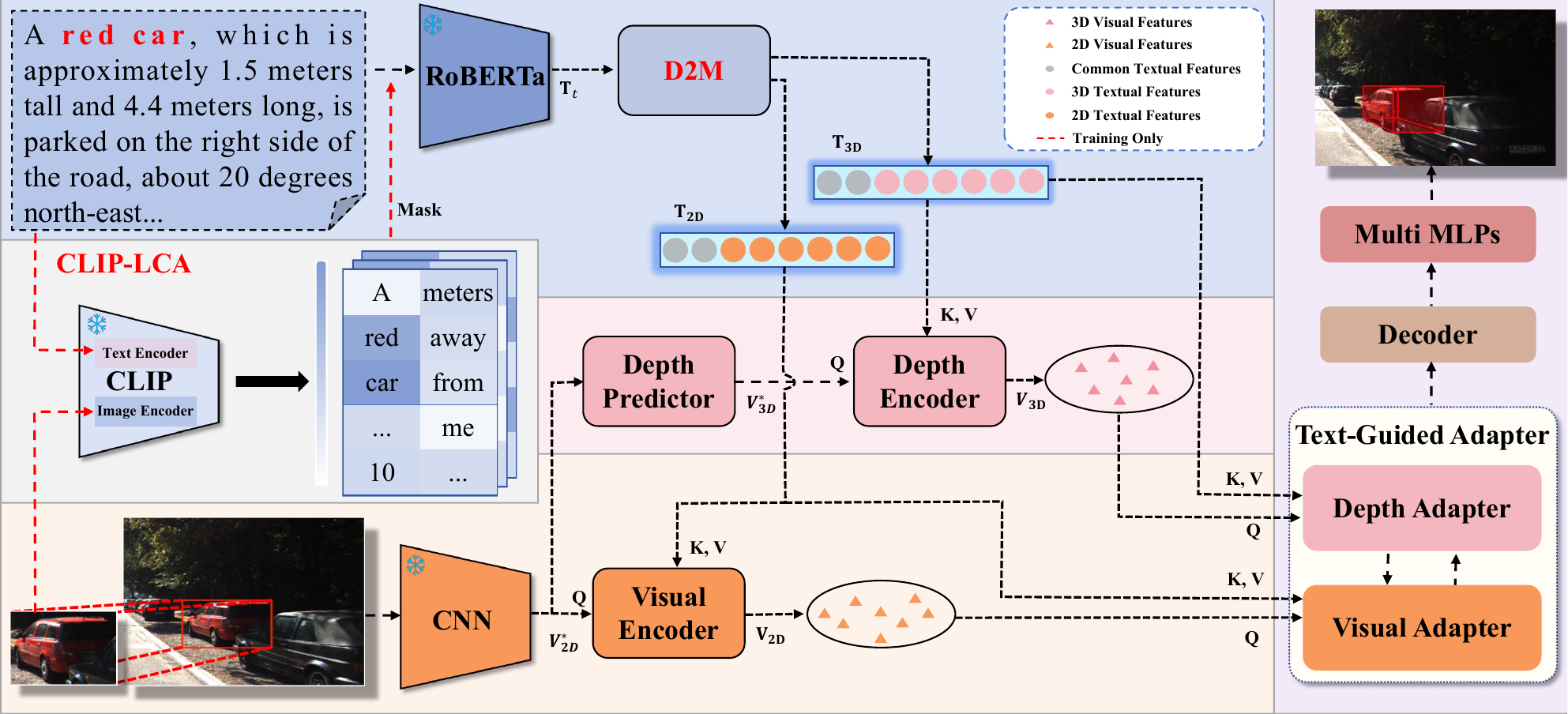} 
\caption{Framework Overview. The architecture integrates multiple feature extraction modules: RoBERTa for generalized textual features ($T_{t}$), visual encoder for 2D visual features ($V_{2D}$), and depth encoder for 3D visual features ($V_{3D}$). CLIP-LCA dynamically adjusts the certainty level of textual descriptions during training. D2M decomposes generalized textual features ($T_{t}$) into 2D-specific ($T_{2D}$) and 3D-specific textual features ($T_{3D}$). The adapter refines dimension-specific features of target objects, followed by decoder and multi-MLPs head for 2D-3D attribute prediction.}
\label{fig2}
\end{figure*}

\section{Method}
Building upon the baseline framework \cite{zhany24}, we begin by reviewing its core architecture as the foundation for our methods. To extract generalized text features ($T_t$), the framework employs RoBERTa-base \cite{liuy19} followed by a linear projection layer. Similarly, multi-scale visual features ($V_{2D}^*$) are extracted using ResNet-50 \cite{hek16} with a linear layer. We follow the method presented in \cite{zhangr22} by implementing a lightweight depth predictor to derive 3D visual features $V_{3D}^*$. We then employ a dual-branch encoder architecture, comprising a visual encoder for 2D feature alignment and a depth encoder for 3D feature interaction. The visual encoder (Eq. 1) leverages a multi-scale deformable attention (MSDA) layer for efficient feature encoding. Subsequently, it incorporates 2D-specific textual information ($T_{2D}$) through a cross-attention module, followed by feature refinement via a feed-forward network (FFN) layer. The depth encoder (Eq. 2) employs a multi-head self-attention layer followed by a feed-forward network (FFN) to encode geometric (3D visual) embeddings, which then interact with 3D-specific textual features ($T_{3D}$) through a cross-attention layer.

\begin{equation}
V_{2D}=FFN(MHCA(MSDA(V_{2D}^*),T_{2D})),
\end{equation}

\begin{equation}
V_{3D}=MHCA(FFN(MHSA(V_{3D}^*)),T_{3D}).
\end{equation}

The adapter module is designed to refine both visual and geometric features. Specifically, the visual features interact with 2D-specific textual features, while the geometric features engage with their 3D-specific textual counterparts. The decoder employs a progressive fusion strategy, sequentially injecting geometric features, generalized textual representations, and 2D visual features into a learnable query via sequential cross-modal attention layers, as detailed in \cite{zhany24}.

\subsection{CLIP-Guided Lexical Certainty Adapter}

To prevent the model from over-relying on explicit lexical cues, CLIP-LCA applies a dynamic strategy to mask high-certainty words during training, as illustrated in Fig.2. Specifically, CLIP-LCA begins by cropping the target region according to the ground-truth annotations and encoding it with the visual branch of CLIP (ViT-B/16) to obtain visual features. Simultaneously, each word in the caption is independently processed by CLIP’s text encoder to generate textual features. Then, we compute the similarity between the textual feature of each word and the target visual feature. Based on these similarity scores, a k-means clustering algorithm (k = 2) is employed to partition the words into high-certainty (high similarity scores) and low-certainty (low similarity scores) categories. The cropped visual content primarily preserves the intrinsic 2D attributes of the object, such as its class name, color, shape. Consequently, words in the textual description corresponding to the object name or color exhibit high similarity with the visual features. In contrast, spatial descriptors that refer to the relative position of the object and its relationship to other objects show low similarity, since the cropping operation removes contextual information about the spatial relationships of the object within the full image. To enhance the text encoder’s understanding of spatial information, we mask words categorized as high-certainty with ``***’’. By masking explicit object cues, this strategy encourages the model to rely on implicit spatial semantics for grounding the corresponding object. Benefiting from this enhanced spatial understanding, the model effectively integrates explicit textual cues (high-certainty words) during inference, enabling more accurate and robust grounding performance.

\subsection{Dimension-Decoupled Module}
To generate dimension-specific text features for dimensionally-consistent refinements, we propose a Dimension-Decoupled module. The detailed architecture of this module is illustrated in Figure 3. The proposed D2M framework consists of two parallel branches: a 2D branch and a 3D branch. Each branch begins with a learnable query that interacts with the generalized textual features $T_{t}$ through a cross-attention layer, followed by a feed-forward neural network (FFN) mapping layer, as formulated below.

\begin{equation}
H_{2D}=MHCA(L_{2D},T_{t}), 
\end{equation}

\begin{equation}
H_{3D}=MHCA(L_{3D},T_{t}),
\end{equation}

\begin{equation}
T_{C2D}=FFN(H_{2D})+H_{2D},
\end{equation}

\begin{equation}
T_{C3D}=FFN(H_{3D})+H_{3D}.
\end{equation}

The learnable queries for two dimensions, denoted as $L_{2D}$ and $L_{3D}$. Through the Eqs. (3)-(6), we derive both 2D and 3D textual features at a coarse level, denoted as $T_{C2D}$ and $T_{C3D}$. To refine the coarse textual features, we propose a reverse cross-attention module, detailed in Equations (7)-(8).

\begin{equation}
T_{2D}\!=\!T_{C2D}\!+\!FFN(Softmax(1\!-\!T_{C2D}\!T_{C3D})T_{C2D}), 
\end{equation}

\begin{equation}
T_{3D}\!=\!T_{C3D}\!+\!FFN(Softmax(1\!-\!T_{C3D}\!T_{C2D})T_{C3D}). 
\end{equation}

The core objective of the reverse cross-attention module is to compute low-attention regions between Query and Key features and subsequently enhance the information corresponding to these low-attention regions. This is achieved by first computing the attention map between the Queries and Keys, then inverting this map by subtracting values from 1, followed by softmax normalization. This inversion shifts attention toward originally low-attention parts, thereby enhancing the contribution of these features during subsequent processes with Values. The D2M framework provides two separate textual features, each encoded with either 2D-specific or 3D-specific details. These textual features guide the subsequent refinements of their corresponding visual features, ensuring dimensionally-consistent cross-modal refinement.

\begin{figure}[t]
\centering
\includegraphics[width=0.9\columnwidth]{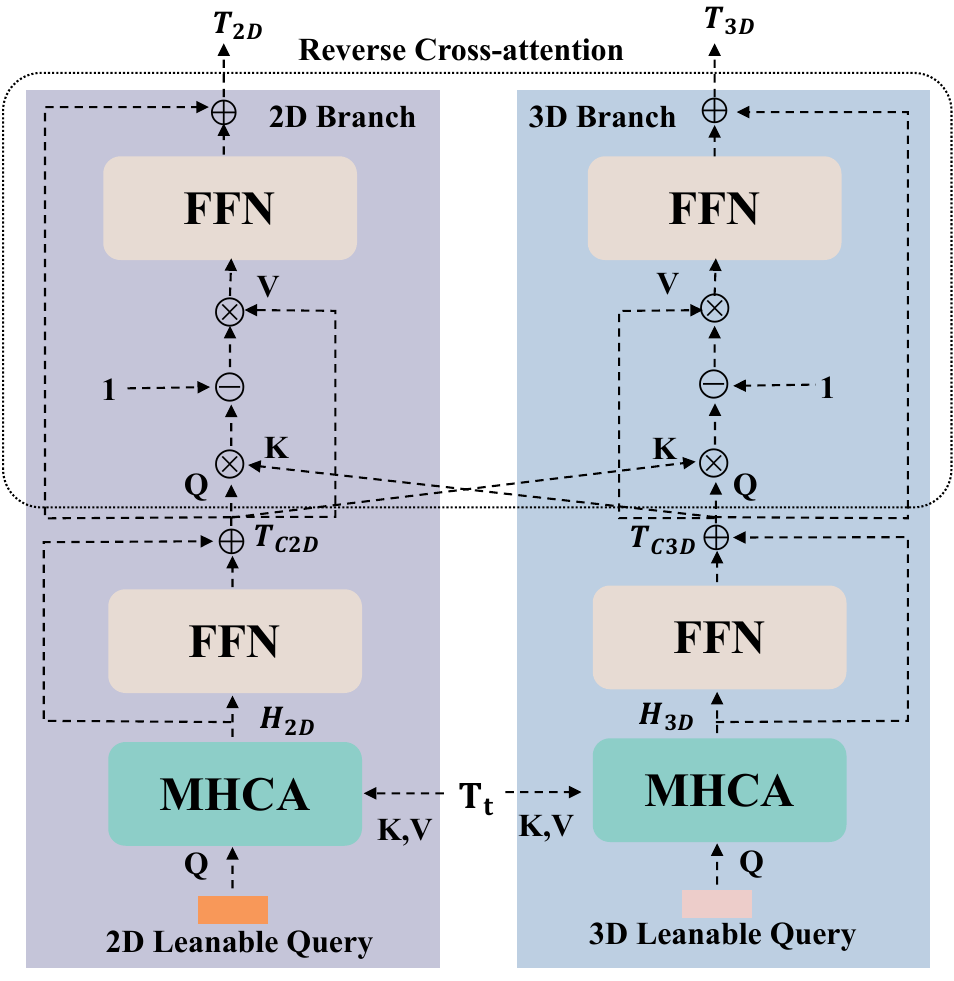} 
\caption{The detailed architecture of Dimension-Decoupled Module. The left part illustrates the procedure for decoupling 2D-specific textual features $T_{2D}$, while the right section presents the corresponding workflow for 3D-specific textual features $T_{3D}$.}
\label{fig3}
\end{figure}

\subsection{Loss Function}
The prediction head utilizes multiple MLPs to estimate both two-dimensional and three-dimensional attributes. The 2D prediction branch outputs classification, 2D box size, and the projected 3D center. For 3D prediction, it outputs 3D box size, orientation, and depth values. The loss function for 2D predictions is formulated as:

\begin{equation}
L_{2D}=\lambda_1L_{classs}+\lambda_2L_{lrtb}+\lambda_3L_{GIoU}+\lambda_4L_{xy3D},
\end{equation}
where $\lambda_{1-4}$ are set to (2,5,2,10) following MonoDETR \cite{zhangr22}. The classification loss $L_{class}$ employs the Focal loss function \cite{linty17} for nine classes prediction. $L_{lrtb}$  and $L_{xy3D}$ utilize the L1 loss function. The 2D bounding box regression is optimized using the GIoU loss \cite{rezatofighih19}, denoted as $L_{GIoU}$. The loss for 3D part is expressed as:

\begin{equation}
L_{3D}=L_{size3D}+L_{orien}+L_{depth},
\end{equation}
where $L_{size3D}$, $L_{orien}$ and $L_{depth}$ represent the 3D IoU oriented loss \cite{max21}, MultiBin loss and Laplacian aleatoric uncertainty loss \cite{chendz20}. To supervise the depth map prediction, we employ a Focal loss, represented as $L_{dmap}$. The total loss function is then defined as follows:

\begin{equation}
L_{overall}=L_{2D}+L_{3D}+L_{dmap}.
\end{equation}

\section{Experiments}

\begin{table*}
  \caption{Comparison of model performance across Unique, Multiple, and Overall scenarios. Underlined results indicate better performance than our bolded accuracy.}
  \label{tab:score1}
  \tabcolsep=0.014\linewidth
  \begin{tabular*}{\linewidth}{cccccccc}
    \Xhline{1pt}
    \multirow{2}{*}{Method}&\multirow{2}{*}{Type}&\multicolumn{2}{c}{Unique}&\multicolumn{2}{c}{Multiple}&\multicolumn{2}{c}{Overall}\\
    &&Acc@0.25&Acc@0.5&Acc@0.25&Acc@0.5&Acc@0.25&Acc@0.5\\
    \Xhline{0.5pt}
    CatRand	&Two-Stage&	\underline{100}&	\underline{100}	&24.47	&24.43	&38.69&	38.67 \\
    Cube R-CNN + Rand	&Two-Stage	&32.76	&14.61	&13.36	&7.21&	17.02&	8.60 \\
    Cube R-CNN + Best	&Two-Stage&	35.29	&16.67&	60.52	&32.99&	55.77	&29.92\\
    \Xhline{0.5pt}
    ZSGNet + backproj	&One-stage&	9.02&	0.29&	16.56	&2.23	&15.14&	1.87\\
    FAOA + backproj	&One-stage&	11.96	&2.06	&13.79	&2.12	&13.44&	2.11\\
    ReSC + backproj	&One-stage&	11.96&0.49&	23.69	&3.94&	21.48	&3.29\\
    \Xhline{0.5pt}
    TransVG + backproj&	Tran.-based&	15.78	&4.02	&21.84&	4.16	&20.70	&4.14\\
    Mono3DVG-TR	&Tran.-based&	57.65&	33.04	&65.92&	46.85	&64.36	&44.25\\
    \Xhline{0.5pt}
    \textbf{Mono3DVG-EnSD(Ours)}	&Tran.-based	&\textbf{66.67}&	\textbf{42.65}&	\textbf{70.17}&	\textbf{55.22}&	\textbf{69.51}	&\textbf{52.85}\\
  \Xhline{1pt}
\end{tabular*}
\end{table*}

\begin{table*}
  \caption{Performance comparisons for near-medium-far scenarios and easy-moderate-hard scenarios. Underlined results indicate better performance than our bolded accuracy.}
  \label{tab:score2}
  \tabcolsep=0.0085\linewidth
  \resizebox{\textwidth}{!}{
  \begin{tabular*}{\linewidth}{cccccccc}
    \Xhline{1pt}
    \multirow{2}{*}{Method}&\multirow{2}{*}{Type}&\multicolumn{2}{c}{Near/Easy}&\multicolumn{2}{c}{Medium/Moderate}&\multicolumn{2}{c}{Far/Hard}\\    &&Acc@0.25&Acc@0.5&Acc@0.25&Acc@0.5&Acc@0.25&Acc@0.5\\
    \Xhline{0.5pt}
    CatRand	&Two-Stage&	31.16/47.29	&31.05/47.26	&35.49/33.92	&35.49/33.92	&52.11/30.83	&\underline{52.11}/30.74 \\
    Cube R-CNN + Rand	&Two-Stage	&17.40/21.12	&11.45/11.41	&18.01/17.85	&8.15/8.01	&14.91/10.56	&6.38/5.18 \\
    Cube R-CNN + Best	&Two-Stage&	67.76/59.66&	41.45/33.05	&60.69/60.56&	30.35/33.45	&34.72/46.25	&17.01/22.52\\
    \Xhline{0.5pt}
    ZSGNet + backproj	&One-stage&	24.87/21.33&	0.59/3.35&	16.74/13.87	&3.71/0.63	&2.15/7.57	&0.07/0.84\\
    FAOA + backproj	&One-stage&	18.03/17.51	&0.53/3.43	&15.64/12.18&	3.95/1.34	&4.86/8.83&	0.62/0.90\\
    ReSC + backproj	&One-stage&	33.68/27.90&	0.59/5.71&	24.03/19.23&	6.15/1.97&4.24/14.41&	1.25/1.02\\
    \Xhline{0.5pt}
    TransVG + backproj&	Tran.-based&	29.34/28.88	&0.86/6.95	&25.05/16.41	&8.02/2.75&	4.17/12.91	&0.97/1.38\\
    Mono3DVG-TR	&Tran.-based&	64.74/72.36&	53.49/51.80	&75.44/69.23&	55.48/48.66&	45.07/49.01&	15.35/29.91\\
    \Xhline{0.5pt}
    \textbf{Mono3DVG-EnSD(Ours)}	&Tran.-based	&\textbf{67.80}/\textbf{80.47}&	\textbf{59.21}/\textbf{62.10}&	\textbf{79.71}/\textbf{71.06}&	\textbf{62.97}/\textbf{55.77}&	\textbf{54.31}/\textbf{52.85}	&\textbf{28.89}/\textbf{37.42}\\
  \Xhline{1pt}
\end{tabular*}
}
\end{table*}
\subsection{Mono3DRefer Dataset}
For our experiments, we adopt the Mono3DRefer dataset \cite{zhany24}, which includes 2,025 images sampled from the KITTI dataset \cite{geiger2012we}. These images are annotated with 41,140 captions with a vocabulary of 5,271 words. In terms of description quantity, Mono3DRefer provides a comparable scale of descriptions as ScanRefer and Nr3D, excluding the template-generated Sr3D dataset. More importantly, Mono3DRefer supports object annotations at significantly longer distances, with a maximum range of 102 meters. In contrast, RGB-D sensors typically operate within 10 meters, while LiDAR systems are generally constrained to around 30 meters.

\subsection{Implementation Details}
We conduct all experiments on an NVIDIA GeForce RTX 3090 GPU, training the model for 60 epochs with the AdamW optimizer under the following configuration: a batch size of 10, initial learning rate $ 10^{-4}$, weight decay of $ 10^{-4}$, and a dropout rate of 0.1. For evaluation, we adopt the 3D IoU thresholds of 0.25 and 0.5, following previous research \cite{linz25}, to assess accuracy across 9 scenarios.

To ensure fair comparisons, we employ the same baseline model as Mono3DVG \cite{zhany24}. \textbf{Two-stage Methods Evaluation}: (1) CatRand randomly selects category-matched ground truth boxes as predictions. (2) (Cube RCNN \cite{brazilg23} + Rand) randomly samples prediction from proposal generated by Cube RCNN. (3) (Cube RCNN \cite{brazilg23} + Best) selects the optimal box match for each proposal to quantify the upper-bound performance of two-stage methods. \textbf{One-stage methods}: We adapt four state-of-the-art 2D visual grounding methods (ZSGNet \cite{sadhua19}, FAOA \cite{zhenyuanyang19}, ReSC \cite{yangs20}, TransVG \cite{dengj21}) to 3D area via back-projection for comparative evaluation. We analyze methods performance across three dimensions: (1) We evaluate the approach under `\textbf{unique}’ and `\textbf{multiple}’ scenarios. The unique scenario involves single-object cases, while the multiple scenario contains multiple objects sharing the same category label; (2) To estimate performance across different distance ranges, we categorize evaluation metrics into three depth ranges: \textbf{Near} (0-15m), \textbf{Medium} (15-35m) and \textbf{Far} (\textgreater35m). The ``near'' scenario presents fewer challenges due to clearer observations, while the ``far'' often involves higher ambiguity. The medium range serves as an intermediate zone that represents the practical scenario in real-world applications; (3) To evaluate the impact of visual occlusion and truncation on performance, we categorize the cases into three levels based on occlusion degree and truncation ratio: \textbf{Easy} cases with no occlusion and truncation ratio below 0.15, \textbf{Moderate} cases including no/partial occluded objects and truncation ratio between 0.15-0.3, and \textbf{Hard} containing severely occluded objects or truncation ratio exceeding 0.3.

\begin{figure*}[htbp]
  \center{\includegraphics[width=\textwidth]{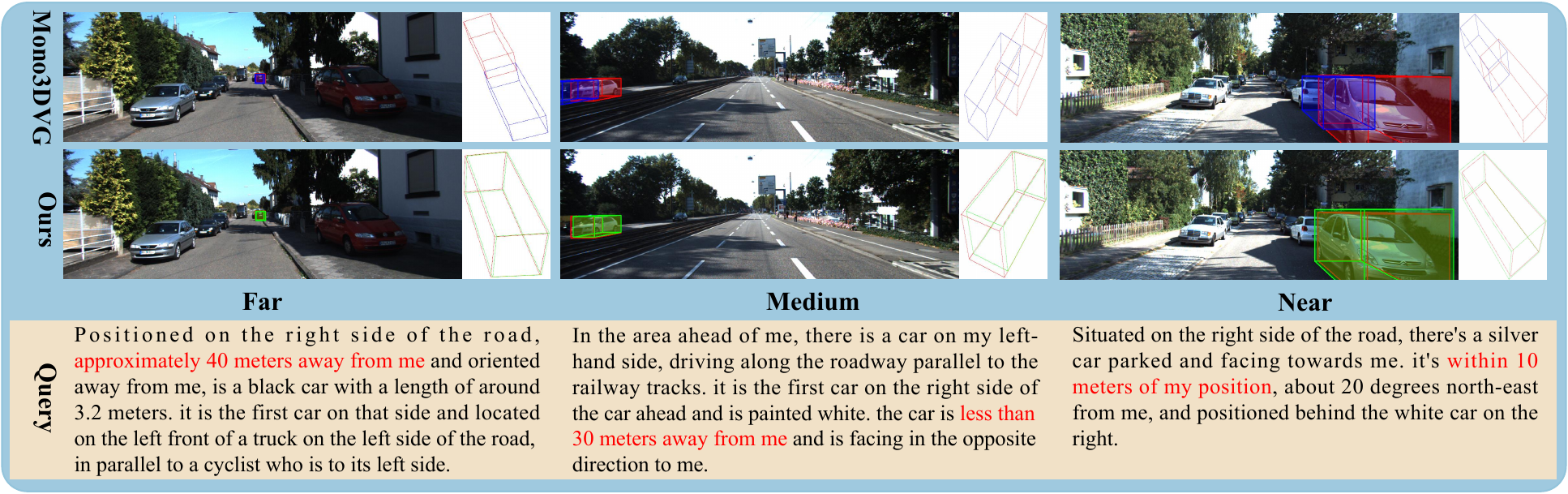}}
  \caption{Visualization of 3D bounding box predictions from Mono3DVG-TR and our Mono3DVG-EnSD. Ground truth (red), our predictions (green), and Mono3DVG-TR predictions (blue) are shown for comparison.}
  \label{fig:4}
\end{figure*}

\subsection{Quantitative Comparisons and Analyses}
As shown in Table~\ref{tab:score1}, CatRand achieves 100\% accuracy in the ``unique'' scenario but drops to 24\% in the ``multiple'' scenario. Similarly, Cube R-CNN Rand performs better on ``unique'' than on ``multiple'' scenarios. When an image contains only a single object, providing the category label is sufficient. However, for images with multiple objects, additional information beyond the label is required to eliminate localization ambiguities, such as spatial or distinct descriptions. As presented in Table~\ref{tab:score2}, the CatRand method demonstrates better performance in the 'far' scenario compared to other distance range scenarios. However, both our method and baseline approaches exhibit progressively lower accuracy as depth increases. The ``far'' scenario contains fewer ambiguous instances, enabling CatRand's random ground truth selection to achieve better performance. In contrast, other methods highly depend on predicted bounding box accuracy, which often leads to imprecise depth and 3D extent, particularly for distinct objects. Our proposed model achieves superior performance in nearly all distance-based scenarios, except for a slight performance gap under the CatRand sampling condition on the ``far'' scenario. Experimental results on the easy-moderate-hard metrics show that our framework achieves state-of-the-art performance. In summary, our method achieves consistent accuracy improvements across all evaluation scenarios. Most notably, under the strict acc@0.5 criterion, we observe substantial gains of +13.54\% for ``far'' scenarios and +10.30\% for ``easy'' scenarios. Meanwhile, our method achieves performance improvements of 9.61\%, 8.37\%, and 8.60\% in ``unique'', ``multiple'', and ``overall'' scenarios (acc@0.5).

\subsection{Qualitative Analysis}
Figure~\ref{fig:4} provides a qualitative comparison of 3D visual grounding results between the baseline Mono3DVG-TR and our proposed Mono3DVG-EnSD, demonstrating the effectiveness of our approach. For clear visual analysis, we select no occlusion and fully visible vehicle samples on three distance intervals: near (\textless15m), medium (15-35m), and far (\textgreater35m) scenarios. The qualitative comparison reveals that while both methods perform competitively in estimating orientation and predicting 3D object dimensions, our method shows superior performance in depth positioning accuracy. The experimental results validate that our Dimension-Decoupled module effectively maintains dimensional consistency during text-visual feature interaction, thereby enabling more accurate refinement of 3D visual features.

\subsection{Ablation Studies}
Ablation experiments were conducted on the Mono3DRefer dataset to assess the proposed CLIP-Guided Lexical Certainty Adapter (CLIP-LCA) and Dimension-Decoupled Module (D2M) component. Their performance is quantified using the standard Acc@0.25 and Acc@0.5 metrics on the ``Overall'' scenario, as summarized in Table~\ref{tab:ablation}. Row 1 corresponds to the baseline model's performance. Row 2 reflects results employing exclusively the CLIP-LCA method, while Row 3 illustrates outcomes integrating only the D2M. Comparison indicates that both components independently yield significant accuracy gains (CLIP-LCA: +2.21\%/+5.04\%; D2M: +3.75\%/+6.83\%), confirming their individual contributions to monocular 3D visual grounding. The final row presents the combined CLIP-LCA+D2M implementation, where synergistic improvement exceeds the individual improvements (+5.15\%/+8.60\%). Due to space limitations, we primarily present the ablation studies of our two key modules in the ``Overall'' scenario. It is worth noting that our modules consistently demonstrate accuracy improvements across all other scenarios as well.

\begin{table}
  \centering
  \caption{The ablation studies of our proposed methods on the ``Overall'' scenario.}
  \label{tab:ablation}
  \begin{tabular}{c|c|cc}
    \Xhline{1pt}
    CLIP-LCA&D2M&Acc@0.25&Acc@0.5\\
    \Xhline{0.5pt}
    -&- & 64.36&44.25 \\
    \checkmark &-& 66.57 &49.29\\
    -&\checkmark & 68.11 &51.08\\
    \checkmark&\checkmark & 69.51 &52.85\\
  \Xhline{1pt}
\end{tabular}
\end{table}

\section{Conclusion}
Existing Mono3DVG methods face two major limitations: over-reliance on high-certainty words and cross-dimensional interference between visual and textual features. To address these challenges, we propose Mono3DVG-EnSD, a novel framework designed to enhance dimensional alignment and understanding ability of spatial descriptions. The proposed CLIP-LCA dynamically suppresses high-certainty words to enhance the model’s comprehension of implicit spatial cues, while the D2M decouples generalized textual features into 2D- and 3D-specific textual features, ensuring dimensionally consistent cross-modal interaction. Our experimental results demonstrate that Mono3DVG-EnSD achieves performance improvements across all nine scenarios compared to the baseline. Through ablation experiments, we further confirm that both the CLIP-LCA and the D2M independently contribute to these performance gains.

\bibliography{aaai2026}

\end{document}